\documentclass[10pt,twocolumn,letterpaper]{article}

\usepackage{wacv}
\usepackage{times}
\usepackage{epsfig}
\usepackage{subfigure} 
\usepackage{graphicx}
\usepackage{amsmath}
\usepackage{amssymb}
\usepackage{cite}
\usepackage{enumitem}
\usepackage{multirow}
\usepackage[table,xcdraw]{xcolor}
\wacvfinalcopy % *** Uncomment this line for the final submission

\def\wacvPaperID{108} % *** Enter the wacv Paper ID here

\ifwacvfinal\fi
\begin{document}

%%%%%%%%% TITLE
\title{ Tracking by Prediction: A Deep Generative Model for Mutli-Person localisation and Tracking}

% Authors at the same institution
%\author{First Author \hspace{2cm} Second Author \\
% Authors at different institutions

\author{Tharindu~Fernando \hspace{1cm} Simon Denman \hspace{1cm} Sridha Sridharan \hspace{1cm} Clinton Fookes\\
\\Image and Video Research Laboratory, Queensland University of Technology (QUT), Australia\\
{\tt\small  \{t.warnakulasuriya, s.denman, s.sridharan, c.fookes\}@qut.edu.au}
}

\maketitle
\ifwacvfinal\thispagestyle{empty}\fi

%%%%%%%%% ABSTRACT
\begin{abstract}
Current multi-person localisation and tracking systems have an over reliance on the use of appearance models for target re-identification and almost no approaches employ a complete deep learning solution for both objectives. We present a novel, complete deep learning framework for multi-person localisation and tracking. In this context we first introduce a light weight sequential Generative Adversarial Network architecture for person localisation, which overcomes issues related to occlusions and noisy detections, typically found in a multi person environment. In the proposed tracking framework we build upon recent advances in pedestrian trajectory prediction approaches and propose a novel data association scheme based on predicted trajectories. This removes the need for computationally expensive person re-identification systems based on appearance features and generates human like trajectories with minimal fragmentation. The proposed method is evaluated on multiple public benchmarks including both static and dynamic cameras and is capable of generating outstanding performance, especially among other recently proposed deep neural network based approaches.
\end{abstract}

\vspace{-3mm}
%%%%%%%%% BODY TEXT
\section{Introduction}
Multi-person localisation and tracking is one of the most active research areas in computer vision as it enables a variety of applications including sports analysis \cite{lu2013identification,fernando2015discovering, xing2011multiple}, robot navigation \cite{elfes1989using,ess2008mobile} and autonomous driving \cite{Fernando_2017_ICCV,ess2009improved,petrovskaya2009model}. \par

Despite the impact of deep learning across a multitude of computer vision domains in recent years, within the tracking space it has been applied in a somewhat piecemeal manner, with it often used for only a specific part of the tracking pipeline. For example, techniques such as \cite{Son2017MultiObject, leal2016learning, wang2016joint} use DCNNs to model subject appearance within a probabilistic tracker. We note that to date, complete deep learning solutions for both localisation and tracking have been limited \cite{milan2017online}. We believe this is due to the scarcity of training data which is large enough to train a complete deep neural network based platform, as well as the complex, variable length nature of multi person trajectories. We combine this with a deep tracking framework that utilises Long Short Term Memory networks (LSTMs) to capture pedestrian dynamics in the scene and track objects via predicting it's future trajectory.\par
In this paper we contribute a novel light weight person detection framework based on Generative Adversarial Networks (GANs)\cite{goodfellow2014generative}, which can be easily trained on the limited data available for multi person localisation. We extend the general GAN framework to temporal sequences and render a probability map for pedestrians in the given sequence. The temporal structure of the proposed GAN allows us to identify pedestrians more effectively, regardless of the motion of other foreground objects in the scene. \par
As illustrated in \cite{milan2016multi}, multi person tracking consists of two subproblems: data association (i.e assigning a unique identifier to the corresponding targets) and inferring the trajectories of the targets. In most data association paradigms the researchers utilise an appearance based model \cite{dorai2017multi,sadeghian2017tracking,babaee2017combined,song2016online,tian2016multi} to re-identify the targets in the next frame. Yet in crowded environments with a high likelihood of occlusions, noisy detections, and poor image resolution, appearance models often fail to generate correct identification of the targets. This results in an un-realistic trajectory generation from the tracking process. To counter this problem we propose a novel tracking framework where the object detections in the next frame are associated with targets via considering their predicted short term and long term trajectories. The trajectory prediction method accounts for the motion of the pedestrian as well as the motion of other people in the local neighbourhood which allows the modelling of the interactions among them. This enables intelligence in the data association process generating human like trajectories even in the presence of occlusions and other image artefacts. To achieve this, we build upon recent advances \cite{milan2017online, fernando2017soft} in pedestrian trajectory modelling approaches and propose a method to capture relationships with neighbourhood dynamics as well as the long term dependencies within the scene context. \par

The major contributions of the proposed work can be summarised as follows:
\begin{itemize}
%\item Introduces a novel pedestrian detection platform based on Generative Adversarial Networks (GAN).
%\item A robust light weight algoritm for data association in multi person tracking problems with the aid of trajectory prediction. 
%\item Human like trajectory generation via associating neighbourhood and scene context in the trajectory prediction framework.
%\item Comprehensive evaluation of proposed models on publicly available benchmarks including videos from both static and dynamic cameras. 
%\item Achieve outstanding performance in MOT challenge benchmark datasets, especially among the deep neural network based approaches. 
\item We introduce a novel pedestrian detection platform based on Generative Adversarial Networks (GANs).
\item We develop a robust light weight algorithm for data association in multi person tracking problems with the aid of trajectory prediction.
\item We generate human like trajectory estimates via the association of neighbourhood and scene context in the trajectory prediction framework.
\item We comprehensively evaluate the  proposed models on publicly available benchmarks including videos from both static and dynamic cameras.
\item We achieve outstanding performance in the MOT challenge benchmark datasets, especially among deep neural network based approaches.
\end{itemize}

\vspace{-2mm}
\section{Related work}

\subsection{Pedestrian detection and localisation}

In classical literature, hand engineered feature vectors composed of different pedestrian characteristics are used to train classifiers on image patches. In \cite{klinger2015probabilistic} the authors use a Random Forest classifier trained by boosting where as in \cite{hoiem2008putting} the authors integrate different sources of information (i.e. foreground information, object shape) with probabilistic graphical models. This was further extended in \cite{schindler2010automatic} where the authors incorporate other parameters for detection such as object position, ground plane parameters and confidence of the detection through a Bayesian network. \par 

With the dawn of deep learning hand engineered feature learning has been superseded by automatic feature learning approaches as they can learn more informative, multiple feature hierarchies. In the first attempts to utilise deep methods for pedestrian detection, authors in \cite{ouyang2012discriminative} built upon the classical model of \cite{felzenszwalb2010object} and used a stack of Restricted Boltzmann Machines. This work was further extended by  Ouyang et al. \cite{ouyang2013single} where the authors account for person-to-person relations. In a similar line of work, Sermanet et al. \cite{sermanet2013pedestrian} used a combination of features from the last two layers of a CNN for the detection task. The detection is performed in a sliding window fashion where different scales are used for detecting in different granularities. \par

Most recently, in \cite{hosang2015taking} the authors discuss the importance of the R-CNN pipeline \cite{girshick2014rich} for pedestrian detection. This method scans through image patches at the superpixel level to determine regions of interest for pedestrians before extracting out CNN features. In the next step extracted features are passed to an SVM which classifies the class of each object. Finally it passes through a localisation layer to determine the specific object location. Even though this R-CNN approach renders accurate detections, the networks structure is inherently complex and computationally expensive. It is prohibitive to scan an image at the superpixel level and perform convolution operations within those extracted patches \cite{jiang2016speed}. Single Shot MultiBox Detector (SSD) \cite{liu2016ssd} and You Only Look Once (YOLO) \cite{redmon2016you} models built upon the RCNN model and propose a simpler network architecture with fewer parameters. Yet when processing videos, the usual approach is to process them framewise, completely discarding the temporal relationships that exist between frames, which can be taken into consideration to help overcome clutter and occlusions. 
\par We build upon the recent success of GANs \cite{goodfellow2014generative,our_iccv} and propose a method for pedestrian localisation in video frames. We expand the GAN framework to videos, mapping the temporal relationships between frames. This accounts for occlusions and false detections which can appear due to the motion of non-pedestrian objects in the scene. 

\subsection{Pedestrian tracking}

%\subsubsection{Data association in tracking}
Related literature often handles the data association via recursive Bayesian filters such as Kalman filters \cite{black2002multi}, particle filters \cite{okuma2004boosted} or relying on the appearance based pedestrian characteristics  \cite{song2008vision,wu2007detection,kim2012online} such as position, height etc. Another line of work has emerged where the multi object trajectories are constructed through optimisation strategies \cite{klinger2015probabilistic,milan2016multi,andriyenko2011multi,butt2013multi}. Yet, instead of learning the features for data association in a data driven way, these approaches use hand engineered features, totally relying on the domain knowledge of the composer. \par
Several attempts have been made to introduce deep learning for data association in tracking. Siamese networks \cite{leal2016learning,wang2016joint} and quadruplet networks \cite{Son2017MultiObject} have been introduced to perform data association by considering appearance features, yet haven't been able to generate substantial impact when comparing their performance against probabilistic methods such as \cite{klinger2015probabilistic,milan2016multi} in public benchmarks.  

%\subsubsection{Pedestrian motion modelling}
%One of the most popular pedestrian motion modelling approach is social force models \cite{robicquet2016learning,pellegrini2009you,yamaguchi2011you,choi2010multiple}  . In social force models each pedestrian reacts based on the energy forces generated from the neighbouring pedestrians. Inspired by the crowd simulation technique another line of work called crowd motion pattern model \cite{zhao2012tracking,kratz2010tracking,rodriguez2009tracking,solera2017tracking} was introduced. Despite their popularity, these approaches rely their motion modelling upon few hand engineered features such as collusion avoidance, group motion patterns, instead of learning those behaviours directly from data. \par

With recent advances in sequence modelling with deep learning approaches, a data driven method for crowd motion modelling has been proposed in \cite{alahi2016social}, which was further expanded to capture the entire history from the neighbourhood in \cite{fernando2017soft}. In a different line of work authors in \cite{fernando2017tree} have looked into modelling long term dependencies such as trajectory patterns between sequences for the task of trajectory prediction, rather than using dependencies within the sequence. However, these methods were engineered for long term prediction of trajectories as opposed to tracking. \par
Numerous attempts have been made to transfer these deep pedestrian motion modelling approaches for data association to tracking.  In \cite{sadeghian2017tracking} the authors utilise LSTMs as motion and interaction models, in addition to using an appearance model to cope with the occlusions, noisy detection and appearance changes. They utilise two separate LSTMs as their motion and interaction models. Still their approach needs tedious processing such as occupancy map generation to obtain interaction features. The authors of \cite{milan2017online} proposed an LSTM based end-to-end approach for multi target tracking, including initiation and termination of objects, bounding box prediction and data association. Yet they do not consider interactions among objects and the approach results in erroneous and non realistic trajectory generation.\par
In the proposed model we show how automatic learning of such interactions is possible in a data driven manner. We introduce a coherent architecture for capturing temporal dependencies from multiple information cues derived from the deep trajectory planner in \cite{fernando2017soft}. In addition to the historical behaviour of the pedestrian of interest we capture information from its local neighbourhood as well as contextual information such as intention and group behavioural patterns, which are stored as latent variables in the trajectory planner. We elaborate on how these information streams can be utilised instead of an appearance model when performing data association. 

\vspace{-2mm}
\section{Tracking framework}

\begin{figure}[t]
\begin{center}
   \includegraphics[width=1\linewidth]{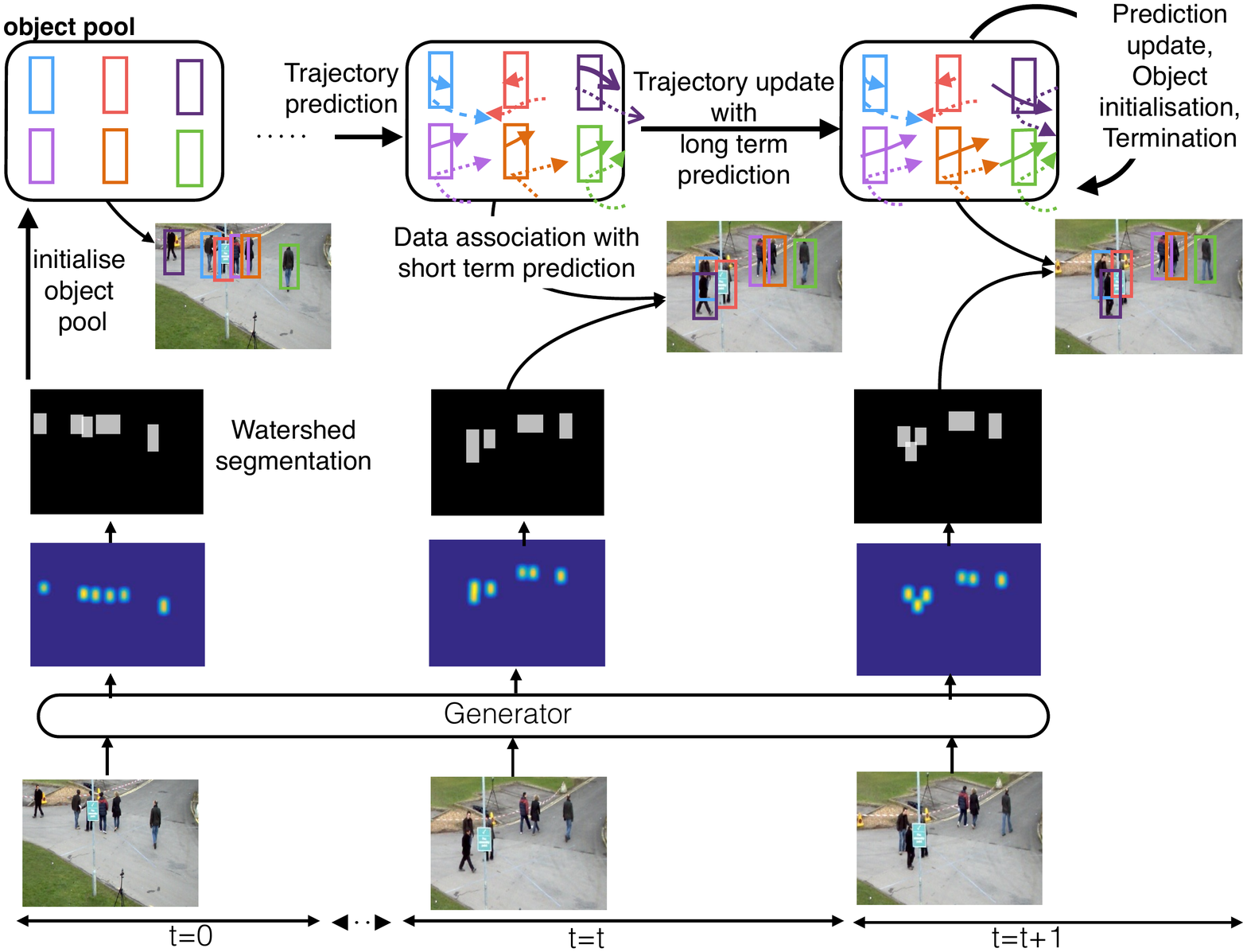}
\end{center}
   \caption{The proposed deep tracking framework: In the first frame of the sequence the object pool is initialised via the person detections generated by passing the input frame through the person detector framework (Section 3.1). We apply watershed segmentation to segment the probability map generated by the generator. Then the trajectory prediction process predicts the short term and long term trajectory predictions where the former is used for data association and the latter is used to update the trajectory of the objects and render human like trajectories in the presence of occlusions and other image artefacts. Finally we update our predictions as well as the object pool by adding newly created objects to the pool and terminating the objects that have not been updated recently.}
\label{fig:tracking}
\vspace{-3mm}
\end{figure}

Fig. \ref{fig:tracking} depicts the proposed tracking framework. We pass each input frame through the GAN generator. This yields a probability map which classifies the likelihood of each pixel of the input frame being a pedestrian or not (see Section 3.1). Then we apply watershed segmentation \cite{shafarenko1997automatic} to segment out each connected component. In order to maintain the trajectory information of each object we retain an object pool in the memory. The segmentation of results of the first frame is used to initialise the object pool. In the proposed framework data association is performed through a trajectory prediction process. The predicted short term trajectory is used for data association where as the long term trajectory prediction is used for trajectory update of the objects to render human like trajectories in the presence of occlusions and other image artefacts. As the final step we update our predictions as well as the object pool by adding newly created objects to the pool and terminating tracking for objects that have not been updated recently. A detailed explanation of the process is given in the following subsections.

\subsection{Person localisation through GANs}
Generative adversarial networks learn a mapping from a random noise vector $z$ to an output image $y, G: z \rightarrow y$ \cite{goodfellow2014generative}. The generator (G) is trained to generate data that is indistinguishable from real data, while the discriminator (D) is trained to identify the generated data. This objective can be written as,
\begin{equation}
\begin{split}
L_{GAN}(G,D)  =E_{x, y \sim  p_{data}(x, y)}[log(D (x, y))] \\
+ E_{x \sim p_{data(x)}, z \sim p_z(z)}[log(1-D(x, G(x,z)))].
\end{split}
\end{equation}

 We utilise a generative model to create a probability map for person detection, which classifies the likelihood of each pixel of the input frame being a pedestrian. We extend the general image to image synthesis framework of GAN to video segments, by incorporating an LSTM module between the encoder and decoder of the generator. This allows us to focus on both spatial and temporal dynamics of the different regions in the scene and generate scene specific detection maps for the pedestrians. \par
Then the loss function of the proposed model for $t$ frames can be written as,
 \begin{equation}
\begin{split}
L_{GAN}(G,D)  =\dfrac{1}{t}\sum_{t=1}^{t} E_{x_t, y_t \sim  p_{data}(x_t, y_t)}[log(D (x_t, y_t))] \\
+ \dfrac{1}{t}\sum_{t=1}^{t} E_{x_t \sim p_{data(x_t)}, z_t \sim p_z(z_t)}[log(1-D(x_t, G(x_t,z_t)))].
\end{split}
\end{equation}

Let $Ck$ denote a Convolution-BatchNorm-ReLU layer group with $k$ filters. $CDk$ denotes a Convolution-BatchNorm-Dropout-ReLU layer with a dropout rate of 50\%. Then the generator architecture can be written as, Encoder: C64-C128-C256-C512-C512-C512-C512-C512 followed by an LSTM module with 64 hidden units. For the decoder we use CD512-CD1024-CD1024-C1024-C1024-C512-C256-C128. The discriminator architecture is: C64-C128-C256-C512-C512-C512 and finally an LSTM with 64 hidden units. The only difference between the LSTM architecture in the encoder and the decoder is that the former is a sequence-to-sequence LSTM which generates an output at each timestep, where as the latter is a sequence-to-1 LSTM which generates a single output considering the entire sequence \cite{gammulle2017two}. All convolutions are 4 x 4 spatial filters applied with stride 2. Convolutions in the encoder and in the discriminator down sample by a factor of 2, whereas in the decoder they up sample by a factor of 2 (i.e fractional Convolutional layers). \par
To sample video segments from the input video we use a sliding window and sample segments with a length of 10 frames. The motivation behind choosing this window length is explained in Sec. \ref{section_tracking}. We trained the proposed model with the Adam \cite{kingma2014adam} optimiser, with a batch size of 32 and an initial learning rate of $1 \times e^{-5}$, for 10 epochs. 

\begin{figure}[t]
\begin{center}
   \includegraphics[width=0.98\linewidth]{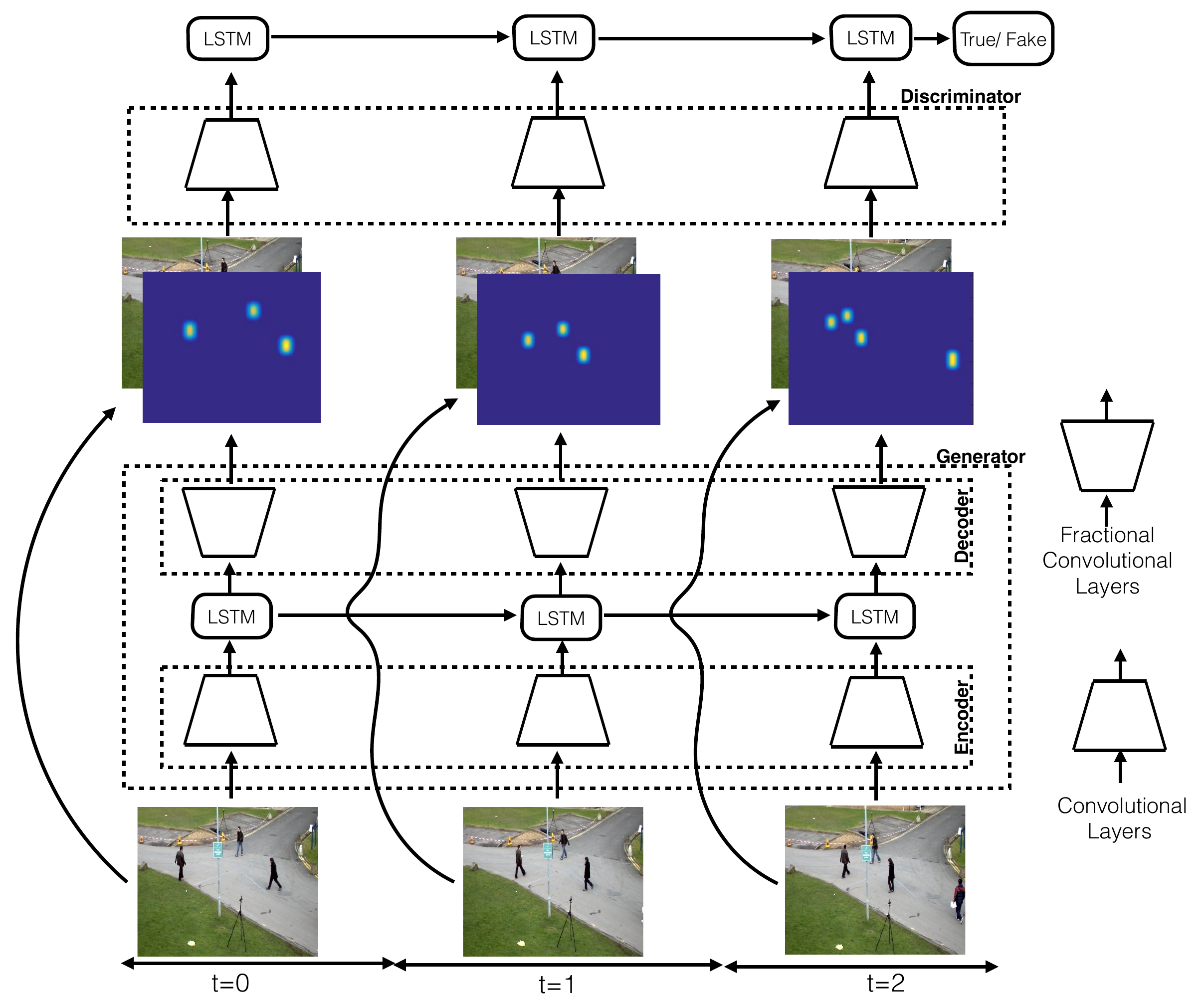}
\end{center}
   \caption{The proposed pedestrian detection framework : Video frames are passed through the encoder of the generative model frame wise. The temporal relationships between the encoder embeddings are mapped through an LSTM layer. Finally the decoder maps these embeddings to a probability distribution which classifies the likelihood of each pixel of the input frame being a pedestrian. The discriminator of the GAN framework takes both the input frame sequence and the generated probability distribution sequence for those frames into consideration and generates a single classification via passing it through an LSTM network.}
\label{fig:gan}
\vspace{-3mm}
\end{figure}

\vspace{-1.5mm}
\subsection{Tracking using Trajectory Prediction}
\label{section_tracking}

\par \textbf{An object pool: }We retain an object pool in the memory which is initialised from the segmentation results of the first frame. The object pool is associated with a trajectory prediction module which regresses short term and long term trajectories for each object in the pool based on the historic trajectory of that object and neighbouring trajectories. Therefore, each object that resides in the object pool is associated with a probable short term and long term trajectory. For the short term trajectory prediction, we utilise a trajectory and local neighbourhood from the previous 2 frames to predict it's trajectory for the next 2 frames. In contrast, for long term prediction we consider the same attributes for the last 10 frames and predict for the next 10 frames. 
\par
The motivation for using both long and short term predictors can be illustrated as follows. In \cite{sadeghian2017tracking} the authors evaluated the occlusion length (in frames) of the MOT validation set and their results show that most occlusions last around 3 frames. Yet there exist around 150 occlusions that last 3-10 frames. These occur largely due to interactions among groups and can be overcome via long term trajectory prediction. Even though we are not using the same dataset, the findings from \cite{sadeghian2017tracking} validates the argument for both short and long term trajectory planning. Furthermore, as pointed out in \cite{sadeghian2017tracking,milan2016multi} raw detection outputs can be noisy due to occlusions, false alarms, inaccurate bounding boxes, and missing detections. A data association process which relies merely on the short term trajectory of an object may be overly sensitive to this noise and be unable to account for important pedestrian trajectory dynamics such as collision avoidance, persistence and group motion.  

\par 
For the trajectory prediction algorithm we build upon the work of \cite{milan2017online} to incorporate the motion of the pedestrian of interest and it's neighbourhood. Authors in \cite{fernando2017soft} show that their algorithm can account for both long and short term trajectory planning including interactions among pedestrians and group motion. Therefore we improve the deep tracking framework in \cite{milan2017online} with the attributes derived in \cite{fernando2017soft} for trajectory prediction and show how those attributes can be utilised in a deep tracking platform, eliminating the need for separate motion and interaction models as in \cite{sadeghian2017tracking}. \par 

\textbf{Trajectory prediction: }
%Fig. \ref{fig:track_prediction} summaries the trajectory prediction algorithm utilised in the proposed work. A sample surveillance scene is shown on the left where the trajectory of the pedestrian of interest is shown in green, and has two neighbours to the left which are shown in purple. He has a single neighbour in the front and there are no neighbours to the right. The neighbourhood encoding scheme is shown on the right. Trajectory information for each pedestrian in the local neighbourhood is encoded with LSTM encoders as follows:
We use the trajectory prediction framework of \cite{fernando2017soft} to predict motion. 
Let the historical trajectory of pedestrian $k$, from frame 1 to frame $T_{obs}$ be given by, 
\begin{equation}
 \mathbf{p^k}=[p_1, \ldots, p_{T_{obs}} ] ,
 \label{eq:input_traj}
\end{equation}
where the trajectory is composed of points in a Cartesian grid. 
Then we pass these historical trajectories through the LSTM encoder of each respective pedestrian to generate its historical embeddings as follows, 
\begin{equation}
h^k_{t}=\mathrm{LSTM}(\mathbf{p}^k_{t},h_{t-1}),
\end{equation}
generating a sequence of historical embeddings,
\begin{equation}
 h^k=[h^k_1, \ldots, h^k_{T_{obs}} ] .
\end{equation}

We utilise a soft attention context vector $C^{s,k}_{t}$ to embed the trajectory information from the pedestrian of interest ($k$), which can be computed as a weighted sum of hidden states, 
\begin{equation}
C^{s,k}_{t}=\sum_{j=1}^{T_{obs}} \alpha_{tj}h^k_j ,
\end{equation}
and the weight $\alpha_{tj}$ can be computed by,
\begin{equation}
\alpha_{tj}=\cfrac{exp(e_{tj})}{\sum_{l=1}^{T} exp(e_{tl})} , 
\end{equation}
\begin{equation}
e_{tj}=a(h^{k}_{t-1},h^k_j) ,
\end{equation}
where the function $a$ is a feed forward neural network for joint training with other components of the system. 

%The hardwired attention context vector $C^{h,k}$ is used for embedding the neighbouring trajectories.  The hardwired weights, denoted by $w$ in Fig. \ref{fig:track_prediction}, can be computed as, 
The hardwired attention context vector $C^{h,k}$ is used for embedding the neighbouring trajectories.  The hardwired weights, denoted by $w$, can be computed as, 
\begin{equation}
	w^n_{j}=\cfrac{1}{\mathrm{dist}(n,j)}  ,
\end{equation}
where $\mathrm{dist}(n,j)$ is the distance between the $n^{th}$ neighbour and the pedestrian of interest at the $j^{th}$ time instant, and $w^n_{j}$ is the generated hardwired attention weight.
When there are $N$ neighbouring trajectories in the local neighbourhood, and $h^{n}_{j}$ is the encoded hidden state of the $n^{th}$ neighbour at the $j^{th}$ time instant, then the context vector for the hardwired attention model is defined as,
\begin{equation}
C^{h,k}=\sum_{n=1}^{N}\sum_{j=1}^{T_{obs}} w^n_{j}h^{n}_{j} .
\end{equation}

The merged context vector, $C_{t}^{*,k}$, computed by, 

\begin{equation}
C_{t}^{*,k}=\mathrm{tanh}([C^{s,k}_{t};C^{h,k}]) ,
\label{eq:context_vector}
\end{equation}

is then used to predict the future trajectory for the pedestrian of interest,

 \begin{equation}
\mathbf{q}_t=\mathrm{LSTM}(h^k_{t-1},\mathbf{q}_{t-1},C_{t}^{*,k}).
\label{eq:tp}
\end{equation}

 %and is shown in red in Fig. \ref{fig:track_prediction}. 
 Note that $\mathbf{q}_t$ is composed of points in a Cartesian gird. We let $T_{obs}=3$ for short term predictions and $T_{obs}=10$ for long term trajectory predictions. After predicting the short term and long term trajectories the predictions are stored in respective objects along with the sequence of context vectors $C^{*,k}$ that is used for long term trajectory prediction. In contrast to \cite{fernando2017soft}, where the authors cluster the trajectories and learn separate motion models, we learn one single motion model to predict all trajectories individually. Similar to \cite{fernando2017soft}, the hidden state dimensions of all the LSTM embeddings (i.e $h^k$) are set to be 300 hidden units. Both short and long term trajectory prediction models are pre-trained on the dataset of \cite{EIF}, with the Adam optimiser and an initial learning rate of $1 \times e^{-4}$ and batch size of 32 for 100 epochs. We fine-tuned on the training set of the respective dataset with an $1 \times e^{-5}$ learning rate, due to limited data availability preventing us training models from scratch. 
 
%\begin{figure}[t]
%\begin{center}
%   \includegraphics[width=0.98\linewidth]{fig/figure_model_learning.pdf}
%\end{center}
%   \caption{Trajectory prediction framework proposed in \cite{fernando2017soft}. A sample surveillance scene (on the left): The trajectory $(\mathbf{p^k}=[p_1, \ldots, p_{T_{obs}} ] )$ of the pedestrian of interest $k$ is shown in green, and has two neighbours (shown in purple) to the left, one in front and none on right. Neighbourhood encoding scheme (on the right): Trajectory information is encoded with LSTM encoders. A soft attention context vector $C^{k,s}_{t}$ is used to embed the trajectory information from the pedestrian of interest, and a hardwired attention context vector $C^{k,h}$ is used for neighbouring trajectories.  In order to generate $C^{k,s}_{t}$ we use a soft attention function denoted $a_t$ in the above figure, and the hardwired weights are denoted by $w^n$. The merged context vector is then used to predict the future trajectory $(p^{k}_{T_{obs+1}}, \ldots, p^{k}_{T_{pred}})$  for the pedestrian of interest (shown in red).}
%\label{fig:track_prediction}
%\end{figure}
\par

\textbf{Data association with short term trajectory predictions: } In the next step we associate each detection (see Section 3.1) with an object from the object pool, if the centroid of the segment lies within a distance threshold to the predicted short term trajectory. The trajectory history of the associated object is updated accordingly.
\par \textbf{New object initialisation: } If no object in the pool lies within the given threshold, a new object is created and added to the pool. \par
\textbf{Trajectory update with long term trajectory prediction:} We utilise a long term prediction module to render smooth human like tracking outputs, overcoming the noisy discontinuous nature of the detections. All the objects that reside in the object pool are compared pairwise and merged if they exhibit a similar long term trajectory and if there is similarity between the context vector sequences of the two pedestrians. 
%In \cite{fernando2017soft} authors show that pedestrian dynamics, contextual information such as intention of the pedestrian, and neighbourhood properties such as walking in a crowd group are captured in the context vector of the model. 

Let the two pedestrians be denoted as $k_1$ and $k_2$ and the long term trajectory predictions for the period $T_{obs+1}$ to $T_{pred}$ (i.e from 11 to 20) be $p^{k_1}=[p^{k_1}_{T_{obs+1}}, \ldots, p^{k_1}_{T_{pred}}]$ and $p^{k_2}=[p^{k_2}_{T_{obs+1}}, \ldots, p^{k_2}_{T_{pred}} ]$ respectively, where $p^k_t$ are points in a 2D Cartesian grid. We denote the context vector sequences from Eq. \ref{eq:context_vector} for the respective pedestrians, for generating the long term trajectory as $C^{*,k_1}=[C_{1}^{*,k_1}, \ldots, C_{T_{obs}}^{*,k_1}]$ and $C^{*,k_2}=[C_{1}^{*,k_2}, \ldots, C_{T_{obs}}^{*,k_2}]$. Then the spatial dissimilarity (SD) between the two objects can measured using the Hausdorff distance \cite{zhang2006comparison} between the long term trajectory predictions of the objects. We use Hausdorff distance as it is widely used to measure the spatial similarity between the pedestrian trajectories in a surveillance setting \cite{lou2002semantic,junejo2004multi}.  This can be denoted as,
\begin{equation}
%TD= \cfrac{\sqrt{\sum\limits_{t=T_{obs}+1}^{T_{pred}}({P^{k_1}_{t}} - P^{k_2}_{t} )^2}}{(T_{pred}-(T_{obs+1}))}.
 SD=\mathrm{max}( d(p^{k_1} , p^{k_2}), d( p^{k_2},p^{k_1} )),
 \label{eq:sd}
\end{equation}
where $d(p^{k_1} , p^{k_2})= \mathrm{max}_{a \in p^{k_1}} \mathrm{min}_{b \in p^{k_2}} ||a-b||$. \par

In \cite{sanborn2015deep} the authors have shown that the cosine-similarity distance measure was more effective for discriminating the hidden states of deep neural networks than traditional methods of using Jacquard similarity or SVMs. Therefore we measure the context dissimilarity (CD),
\begin{equation}
\begin{split}
%CD= \cfrac{\sqrt{\sum\limits_{t=1}^{T_{obs}}({C^{*,k_1}_{t}} - C^{*,k_2}_{t} )^2}}{T_{obs}}.
CD &=1-\cfrac{C^{*,k_1} \cdot C^{*,k_2}}{||C^{*,k_1}||_2 ||C^{*,k_2}||_2}
\\
 & =1- \cfrac{\sum\limits_{t=1}^{T_{obs}}{C^{*,k_1}_{t}}{C^{*,k_2}_{t}}}{\sqrt{\sum\limits_{t=1}^{T_{obs}}({C^{*,k_1}_{t}} )^2}\sqrt{\sum\limits_{t=1}^{T_{obs}}({C^{*,k_2}_{t}} )^2}}
\end{split}
\label{eq:cd}
\end{equation}
Then we retain only the older object, and discard the young object from the object pool if the dissimilarities (i.e $SD$ and $CD$ separately) between the two objects are less than specified thresholds, which are evaluated experimentally and available in the supplementary material. \par
\textbf{Termination: }All the objects that do not get updated for 10 consecutive frames are removed from the pool. 
\par \textbf{Prediction update: }Finally we update the predictions for the next time segment. As depicted in Fig. \ref{fig:tracking}, we repeat this process for all the frames in the given video. 

\vspace{-2mm}
\section{Experimental Results}

\subsection{Person Detector evaluation}

We evaluate the proposed person detector on the PETS2009-S1-L2 \cite{ferryman2009pets2009} dataset which is a widely recognised benchmark dataset for pedestrian detection. It contains seven outdoor sequences
from seven cameras, with 795 frames for every sequence. Similar to \cite{chavdarova2017deep,peng2015robust,ge2010crowd} we retained camera view 1 for testing and trained our detector on the other camera views. In addition to the empirical precision and recall estimates of the detector, we also report the Multiple Object Detection Accuracy (MODA) and Multiple Object Detection Precision (MODP) metrics from \cite{kasturi2009framework}. MODA accounts for the normalised missed detections and MODP assesses the localisation quality of the true positives.

\begin{table}[]
\centering
\caption{Person detection evaluations on PETS S1L2 dataset}
\label{tab:detection}
\resizebox{.98\linewidth}{!}{
\begin{tabular}{|l|l|l|l|l|}
\hline
Method            & MODA          & MODP		 & Precision     & Recall        \\ \hline
Peng et al. \cite{peng2015robust}      & 0.79          & 0.74                      & 0.92          & \textbf{0.91} \\ \hline
Ge et al.  \cite{ge2010crowd}       & 0.75          & 0.68                      & 0.85          & 0.89          \\ \hline
Chavdarova et al. \cite{chavdarova2017deep} & 0.88          & 0.75                      & 0.97          & \textbf{0.91} \\ \hline
Proposed          & \textbf{0.91} & \textbf{0.79}             & \textbf{0.98} & \textbf{0.91} \\ \hline
\end{tabular}
}
\end{table}

From the results shown in Tab. \ref{tab:detection} our method performs well compared to existing state-of-the-art baselines, which is largely due to the augmented temporal structure of the GAN. This provides the detector the capability to leverage appearance information, and filter out foreground objects by considering both appearance and motion features, performing a precise detection of the humans. \par

Furthermore, we evaluate our person detector performance on the PETS S2L1 dataset which is one of the videos available for training in 3D MOT 2015 benchmark \cite{leal2015motchallenge} and compare it against three recent baseline methods. It should be noted that POM-CNN\cite{baque2017deep} utilises a CNN-based foreground segmentation process while DOR\cite{baque2017deep} has coupled the CNN with conditional random fields specifically to handle occlusions and imitate the generative-discriminative training approach of GANs. Still the proposed approach outperforms the state-of-the-art methods in all the considered metrics. We were unable to compare the MODP metric as that information for the baselines is not available. It should be noted that the proposed generative model for pedestrian detection has only 32,869,313 parameters for training where as the region proposal network of the Tiny-Yolo \cite{redmon2016yolo9000} has 45,079,472 parameters. 
\begin{table}[]
\centering
\caption{Person detection evaluations on PETS S2L1 dataset}
\label{my-label}
\begin{tabular}{|l|l|l|l|}
\hline
Method     & MODA          & Precision     & Recall        \\ \hline
DOR \cite{baque2017deep}       & 0.60          & 0.93          & 0.87          \\ \hline
POM-CNN \cite{baque2017deep}    & 0.43          & 0.90          & 0.86          \\ \hline
Faster R-CNN \cite{ren2015faster} & 0.27          & 0.50          & 0.63          \\ \hline
Proposed   & \textbf{0.69} & \textbf{0.95} & \textbf{0.91} \\ \hline
\end{tabular}
\end{table}

\subsection{Tracker evaluations}
 
This section evaluates the proposed deep tracker on the 3D MOT 2015 benchmark \cite{leal2015motchallenge}, which is composed of the PETS09-S2L2 and AVG-TownCentre sequences. For a fair comparison with other baselines we use the provided pedestrian detections. Furthermore, we evaluated the proposed model on ETH Mobile Scene (ETHMS) dataset \cite{ess2008mobile}, which is challenging with a busy pedestrian street filmed with a moving stereo camera. It should be noted that we do not use the available camera calibration  or depth maps, but rather track the people in the image space.  \par

The reported metrics are the ones suggested in the 3D MOT 2015 benchmark: Multiple Object Tracking Accuracy (MOTA), Multiple Object Tracking Precision (MOTP), Mostly Tracked targets (MT), Mostly Lost targets (ML), False Positives (FP), False Negatives (FN), ID Switches (IDS), and the number of frames processed in one second (Hz) denoting the speed of a tracking method.

\begin{table*}[]
\centering
\caption{3D MOT 2015 results for PETS09-S2L2. Arrows indicate favourable direction of each metric. Best values are printed in bold}
\label{tab:PETS09-S2L2}
\resizebox{.98\linewidth}{!}{
\begin{tabular}{|l|l|l|r|l|l|l|l|l|l|l|}
\hline
Type&Tracker                    & MOTA $\uparrow $         & MOTP $\uparrow $                      & MT (\%) $\uparrow $            & ML (\%) $\downarrow$            & FP  $\downarrow$          & FN $\downarrow$            & IDS $\downarrow$          & Frag  $\downarrow$        & Hz $\uparrow $            \\ \hline
\multirow{7}{*}{Probabilistic}  &Klinger et al. \cite{klinger2015probabilistic}      & \textbf{57.6} & 65.5             & \textbf{28.6} & 4.8          & 805          & 3049 & 231          & 245          & 0.1            \\ \cline{2-11}
&Klinger et al. \cite{klinger2017probabilistic}     & 55.5          & 63.6                      & 21.4          & 4.8          & \textbf{638} & 3480          & \textbf{174} & \textbf{202} & 0.1           \\ \cline{2-11}
&Leal- Taixe' et al. \cite{leal2011everybody} & 41.3          & 55.7                      & 7.1           & 16.7         & 640          & 4776          & 243          & 271          & 8.4            \\ \cline{2-11}
&Pellegrini et al. \cite{pellegrini2009you}   & 32.2          & 55.7                      & 4.8           & \textbf{2.4} & 1549         & 4091          & 893          & 889          & 30.6           \\ \cline{2-11}
&MOT baseline  \cite{leal2015motchallenge}             & 45.4          & 54.1                      & 16.7          & 7.1          & 1407         & 3593          & 268          & 296          & 83.5  \\ \cline{2-11}
&Wen et al.  \cite{wen2017multi}          & 47.2          & 56.6		      & 11.9          & 4.8          & 1,140        & 3,710         & 245          & 292          & 1.9            \\ \cline{2-11}
&Milan et al. \cite{milan2016multi}          & 37.5          & 70.7			      & 4.8           & 16.7         & \textbf{638}        & 5,200         & 189          & 209          & 0.3            \\ \cline{2-11} \hline \hline

\multirow{7}{*}{Deep Neural Network}  &Milan et al. \cite{milan2017online}          & 38.3          & 71.6		      & 9.5           & 14.3         & 1,016        & 4,611         & 320          & 417          & \textbf{165.2}           \\ \cline{2-11}
&Sadeghian et al. \cite{sadeghian2017tracking}	&47.0 & 70.5 		     & 11.9         & 9.5 		& 616     & 4,236 	  & 254 	    & 397 		&1.9 \\ \cline{2-11} 
&Bae et al. \cite{bae2017confidence} 			&42.5  &69.3	             &7.1 	         &7.1 	&934	          &4,409	    &196	        &438        &  2.3	 \\ \cline{2-11}
&Wang et al. \cite{wang2016joint}			&49.6   &70.7		     &11.9 	        &11.9 	&780	  	  &3,886	    &192		&218		&1.7	 \\ \cline{2-11}
&Leal-Taixé et al. \cite{leal2016learning}		&34.5	&69.7		&7.1 		&19.0 	&672	 	  &5,364	    &282		&424		&52.8	 \\ \cline{2-11}
&Son et al. \cite{Son2017MultiObject}			&49.0	&72.6		&16.7 	&7.1 		&686		  &3,947	    &285		&380		&3.7 	 \\ \cline{2-11} 
& \cellcolor[HTML]{C0C0C0}Proposed                   &\cellcolor[HTML]{C0C0C0} \textbf{57.6 }             & \cellcolor[HTML]{C0C0C0}\textbf{72.8}   & \cellcolor[HTML]{C0C0C0}\textbf{28.6}               & \cellcolor[HTML]{C0C0C0} 4.7           &  \cellcolor[HTML]{C0C0C0}802            & \cellcolor[HTML]{C0C0C0}\textbf{3043}              & \cellcolor[HTML]{C0C0C0} 224            &  \cellcolor[HTML]{C0C0C0}212            & \cellcolor[HTML]{C0C0C0}78.4    \\ \hline

\end{tabular}
}
\end{table*}

\begin{table*}[]
\centering
\caption{3D MOT 2015 results for AVG-TownCentre. Arrows indicate favourable direction of each metric. Best values are printed in bold}
\label{tab:AVG-TownCentre}
\resizebox{.98\linewidth}{!}{
\begin{tabular}{|l|l|l|r|l|l|l|l|l|l|l|}
\hline
Type& Tracker                    & MOTA $\uparrow $          & MOTP  $\uparrow $                     & MT (\%) $\uparrow $              & ML (\%) $\downarrow$            & FP $\downarrow$           & FN $\downarrow$           & IDS $\downarrow$         & Frag $\downarrow$        & Hz  $\uparrow $           \\ \hline
\multirow{7}{*}{Probabilistic}  &Klinger et al. \cite{klinger2015probabilistic}      & 42.4 & 57.1             & \textbf{28.8} & 20.4          & 1272          & 2697 & 149          & 173          & 0.1           \\ \cline{2-11}
&Klinger et al. \cite{klinger2017probabilistic}      & 42.2          & 57.4                      & 26.5          & 19.5          & 1175	 & 2820          & 137 & 184 & 0.1          \\ \cline{2-11}
&Leal- Taixe' et al. \cite{leal2011everybody} & 28.7          & 51.9                      & 15.0          & 22.6          & 1391          & 3430          & 277          & 330          & 8.4           \\ \cline{2-11}
&Pellegrini et al. \cite{pellegrini2009you}   & 15.2          & 51.4                      & 7.1           & \textbf{16.8} & 1612          & 3508          & 945          & 797          & 30.6           \\ \cline{2-11}
&MOT baseline    \cite{leal2015motchallenge}           & 23.2          & 52.2                      & 21.7          & 18.1          & 2181          & 3000          & 312          & 363          & 83.5  \\ \cline{2-11}
&Wen et al. \cite{wen2017multi}          & 16.8          & 54.2		      & 11.1          & 29.2          & 1,917         & 3,744         & 287          & 319          & 1.9            \\ \cline{2-11}
&Milan et al. \cite{milan2016multi}          & 8.2          & 69.9			      & 2.7           & 69.5         & 763        & 5,766         & \textbf{30}          & \textbf{84}          & 0.3           \\ \cline{2-11}\hline \hline

\multirow{7}{*}{Deep Neural Network}  &Milan et al. \cite{milan2017online}          & 13.4          & 68.8		      & 3.5          & 41.2          & 1,206         & 4,682         & 299          & 414          & \textbf{165.2}          \\ \cline{2-11}
&Sadeghian et al. \cite{sadeghian2017tracking}	&36.2 & 69.5 		      & 26.1 	& 17.7 	   & 1,448 		& 2,882 	    & 234 	      & 389 		& 1.9  \\ \cline{2-11}
&Bae et al. \cite{bae2017confidence} 			&30.7 &68.9		      &13.7  	&31.9 	& 1,013	         &3,807	     &136	      & 367 		&	2.3	   \\ \cline{2-11}
&Wang et al. \cite{wang2016joint}			&31.3  &69.5		     &16.8 	        &33.2 	&952			 &3,825	     &137	       &246		&	1.7	   \\ \cline{2-11}
&Leal-Taixé et al. \cite{leal2016learning}		&19.3	&69.0	    &4.4 		&44.7 	&\textbf{698}			 &4,927	     &142	       &289		&	52.8		 \\ \cline{2-11}
&Son et al. \cite{Son2017MultiObject}			&30.8	&\textbf{69.8}	   &18.1 	        &31.4 	&1,191		&3,643	     &111	       &409		&	3.7 	\\ \cline{2-11}  \cline{2-11}

&\cellcolor[HTML]{C0C0C0}Proposed                   &  \cellcolor[HTML]{C0C0C0}\textbf{42.5}             & \cellcolor[HTML]{C0C0C0} \textbf{69.8}    &  \cellcolor[HTML]{C0C0C0}27.0             &   \cellcolor[HTML]{C0C0C0}19.5            &  \cellcolor[HTML]{C0C0C0}1182             &   \cellcolor[HTML]{C0C0C0}2826            & \cellcolor[HTML]{C0C0C0}139             &   \cellcolor[HTML]{C0C0C0}186           &   \cellcolor[HTML]{C0C0C0}78.4            \\\hline
\end{tabular}
}
\end{table*}
For the results presented in Tab. \ref{tab:PETS09-S2L2} and Tab. \ref{tab:AVG-TownCentre} for the 2 sequences of the 3D MOT 2015 benchmark, it can be observed that the proposed system exhibits better performance among all the models for most metrics. Furthermore, the proposed algorithm exhibits a noticeable increase in both MOTA  and MT, and reductions in ML and trajectory fragmentations (Frag), compared to other deep neural network approaches. Specifically, we compare the proposed method against state-of-the-art deep neural network based methods. Wang et al. \cite{wang2016joint} performs data association using both motion and appearance feature cues. The appearance features are learnt using CNNs where as the motion based track similarity is evaluated based on the velocity, which is handcrafted. Yet their method generates lower MOTA and MT values and very high ML values, in both sequences compared to our method. Furthermore, we would like to point that the CNN based appearance feature extraction method has led their architecture to have a high time complexity. \par
A simpler data association scheme is proposed in  \cite{leal2016learning} where the authors only use appearance features and utilise linear programming to associate the tracklets, which leads to a comparatively higher speed. Even though they are achieving fewer FPs, their tracker misses the majority of the targets resulting a higher FN value and ML percentage, and a lower MT rate. The MOTA value associated with their tracker reflects our observations. \par
Son et al. \cite{Son2017MultiObject}	associates the temporal distance between frame patches in a video with appearance based feature matching using CNNs. They utilise detection properties such as centre position, height, velocity and temporal distance in addition to the convolution features, in their data association framework. Yet their method leads to very high fragmentation of trajectories, frequent id switches and higher FN values while also having a high time complexity (i.e a lower Hz) in both sequences. \par 
Furthermore, we would like to draw comparisons with \cite{milan2017online}, which utilises a deep learning based trajectory planing method for data association and eschews appearance features. Even though with their single LSTM based approach, which does not incorporate any neighbourhood or context information, they were able to obtain a low computational complexity, the method leads to frequent id swapping (i.e higher IDS), higher fragmentation and a very low MT value. Our proposed approach, while similar in that we rely on motion prediction only, has an increased capacity through incorporating the complete history of the neighbourhood as well as the contextual factors derived directly from trajectory modelling. Thus the proposed model has been able to generate results with higher performance. \par 

\textbf{Impact of using multiple trajectory predictions.} We investigate the contributions of each prediction component in our framework for the data association task by measuring the performance of using each component separately and together in terms of tracking performance, in the training set of 3D MOT 2015 benchmark. A detailed evaluation of each tracking method against various MOT matrices is presented in Tab. \ref{tab:albation_study}. The details on each system are as follows:
\begin{enumerate}[label=\textbf{T.\arabic*}]
\item  \label{T1} System with only data association with short term trajectory prediction (STP) (i.e Eq. \ref{eq:tp})
\item  \label{T2} System with STP and trajectory update with only spatial dissimilarity (i.e Eq. \ref{eq:tp} + Eq. \ref{eq:sd})
\item  \label{T3} System with STP and trajectory update with only context dissimilarity (i.e Eq. \ref{eq:tp} + Eq. \ref{eq:cd})
\item  \label{T4} System with STP and trajectory update with combined dissimilarity (i.e Eq. \ref{eq:tp} + Eq. \ref{eq:sd} + Eq. \ref{eq:cd})
\end{enumerate}
The predicted short term trajectory is the central driving mechanism in the proposed framework, due to the fact that most interactions and occlusions occur over short periods of time. It should be pointed out that each component (SD and CD) of the proposed trajectory update mechanism positively impacts on the overall performance as it lowers the possibility of ID switches and trajectory fragmentation. The results on the trajectory update process with the combination of the dissimilarity measures implies that exploiting the data association with the proposed update process can significantly improve the performance of the tracker. 
%\begin{figure}[t]
%\begin{center}
%   \includegraphics[width=0.98\linewidth]{fig/ablation_study.pdf}
%\end{center}
%   \caption{Analysis of the contribution of various components (\ref{T1}, \ref{T2}, \ref{T3}, \ref{T4} ) of our tracking framework. We report the MOTA, MOTP and MT scores on the MOT training set.}
%\label{fig:ablation_study}
%\end{figure}

\begin{table}[]
\centering
\caption{Analysis of the contribution of each component of the proposed tracking framework on the training set of the MOT benchmark. Arrows indicate favourable direction of each metric. Best values are printed in bold}
\label{tab:albation_study}
\resizebox{.98\linewidth}{!}{
\begin{tabular}{|l|l|r|l|l|l|l|l|l|l|}
\hline

Tracker                    & MOTA $\uparrow $          & MOTP $\uparrow $                      & MT (\%) $\uparrow $             & ML (\%)  $\downarrow$           & FP $\downarrow$           & FN  $\downarrow$          & IDS $\downarrow$          & Frag $\downarrow$                 \\ \hline
 T1     			& 42.3           & 48.2             		& 26.7		      & 5.5          & 1032            & 2687          &  329         &  330                   \\ \hline
T2      			& 51.4          & 56.3                      	& 27.3          	      & 4.5          & 877	       & 2520          & 207          & 184 		           \\ \hline
T3      			& 51.4          & \textbf{65.9}                      	& 28.2                     & \textbf{3.7}          & 892           & 2430          & 177          &    191              \\ \hline
T4     			& \textbf{57.9}          & \textbf{65.9}                     	 & \textbf{31.5}                     & 3.6            & \textbf{506}          & \textbf{2187}          & \textbf{146}          & \textbf{105}                  \\ \hline
\end{tabular}
}
\end{table}

\begin{figure}[!h]
\begin{center}
\subfigure[\cite{milan2016multi}, frame 117]{\includegraphics[width = .3 \linewidth]{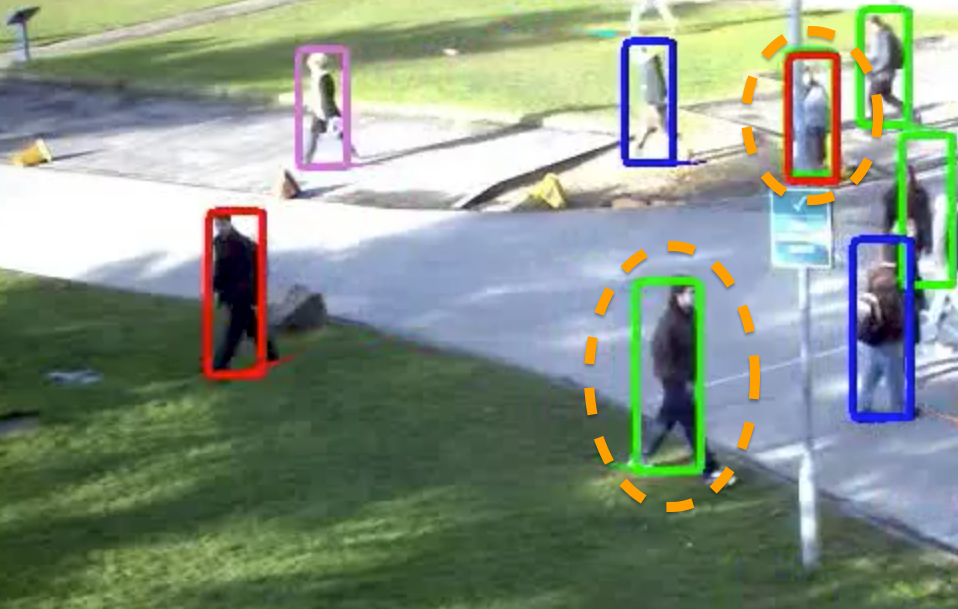}}
\subfigure[\cite{milan2016multi}, frame 125]{\includegraphics[width = .3 \linewidth]{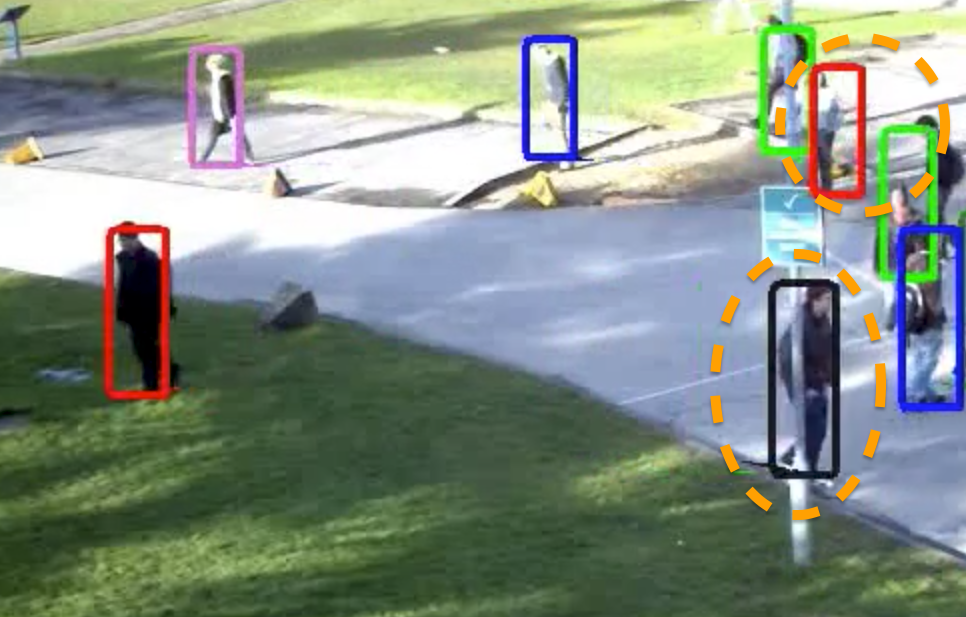}}
\subfigure[\cite{milan2016multi}, frame 128]{\includegraphics[width = .3 \linewidth]{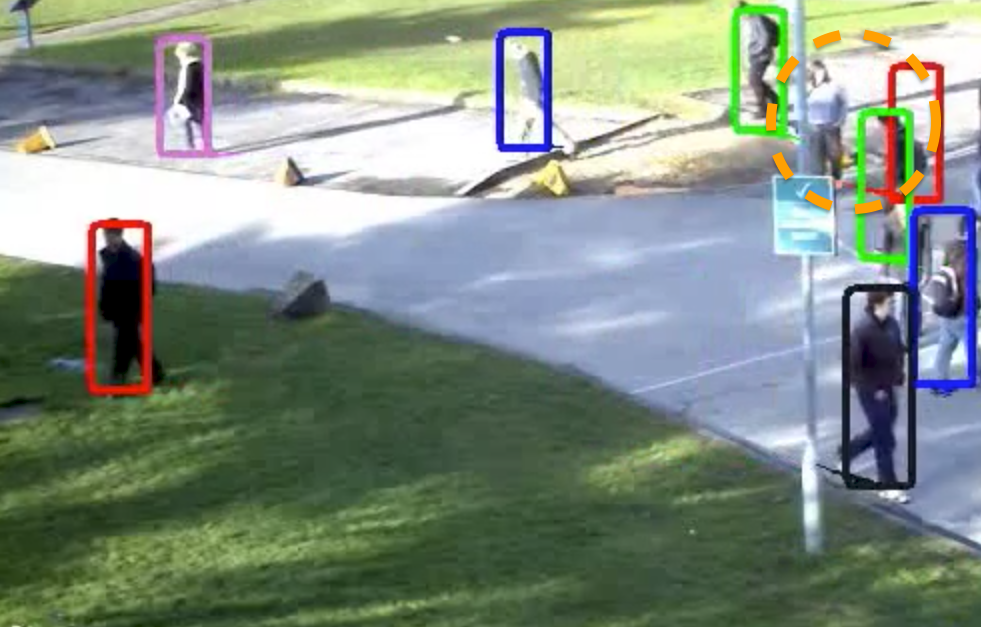}}
\vspace{-1.5mm}
\subfigure[\cite{milan2017online}, frame 117]{\includegraphics[width = .3 \linewidth]{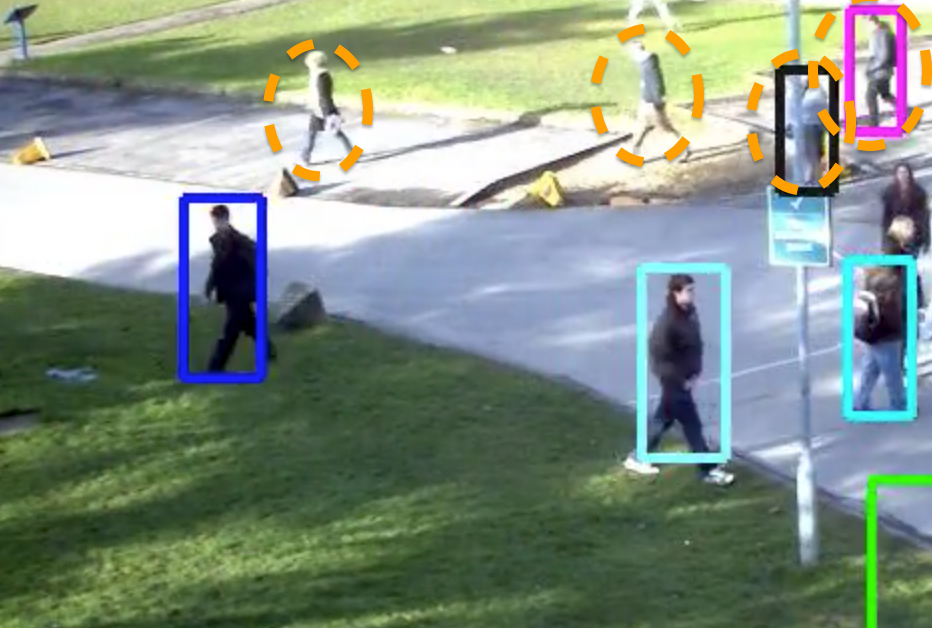}}
\subfigure[\cite{milan2017online}, frame 125]{\includegraphics[width = .3 \linewidth]{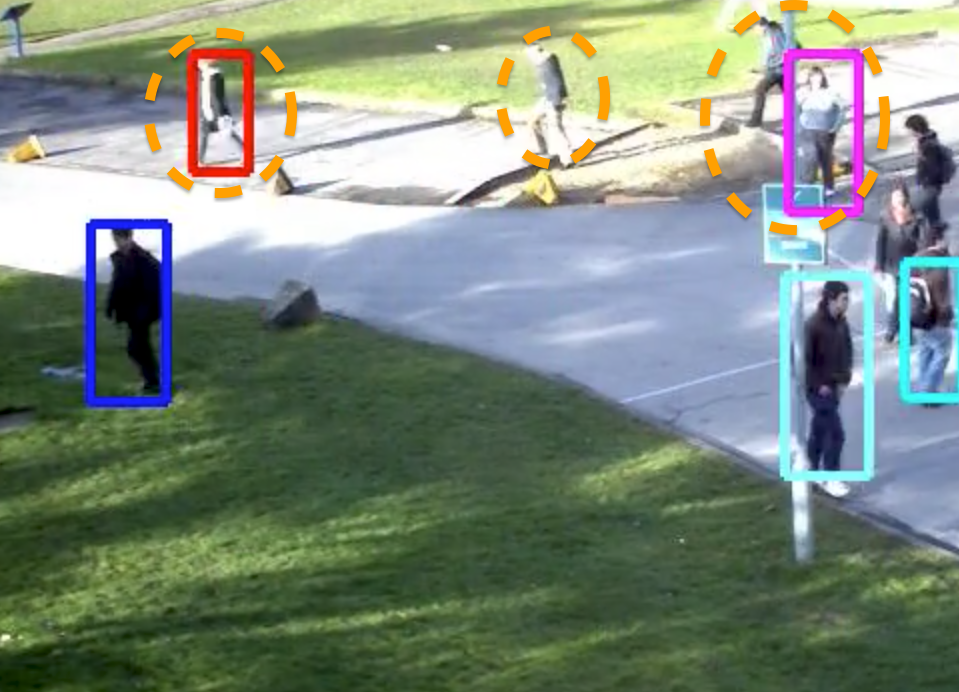}}
\subfigure[\cite{milan2017online}, frame 128]{\includegraphics[width = .3 \linewidth]{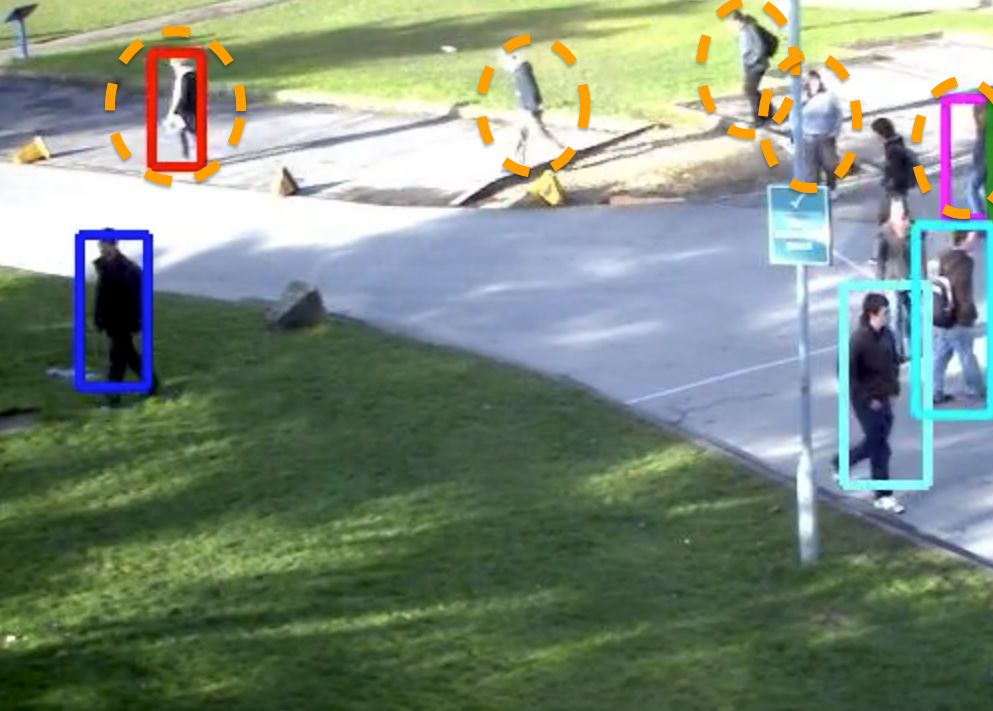}}
\vspace{-1.5mm}
\subfigure[\cite{klinger2015probabilistic}, frame 117]{\includegraphics[width = .3 \linewidth]{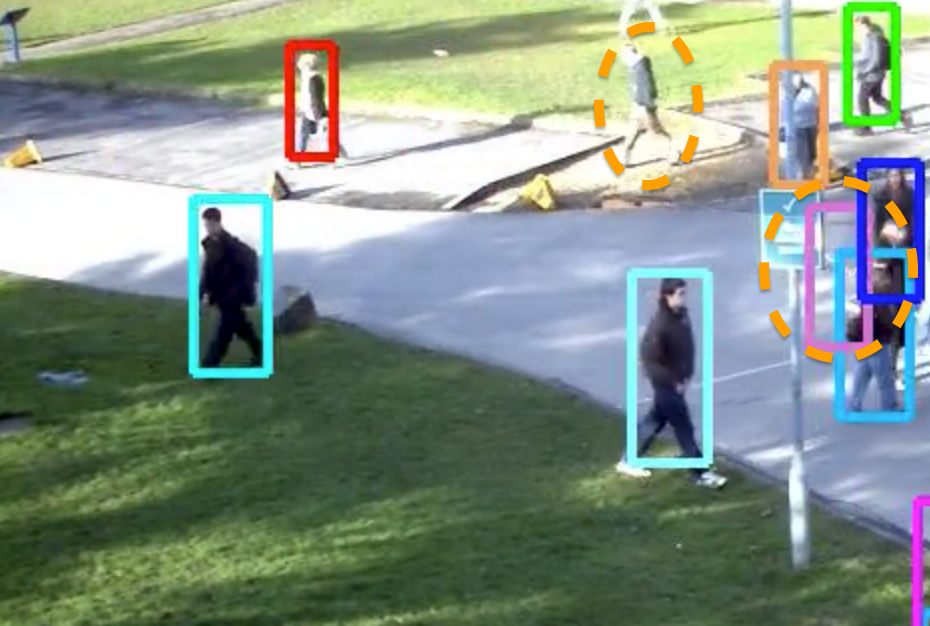}}
\subfigure[\cite{klinger2015probabilistic}, frame 125]{\includegraphics[width = .3 \linewidth]{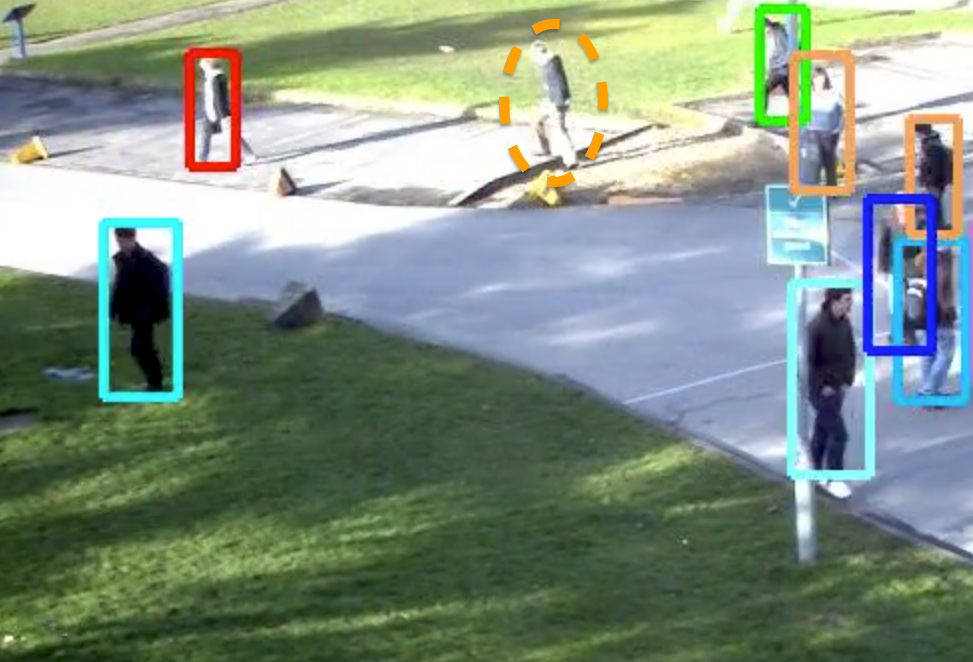}}
\subfigure[\cite{klinger2015probabilistic}, frame 128]{\includegraphics[width = .3 \linewidth]{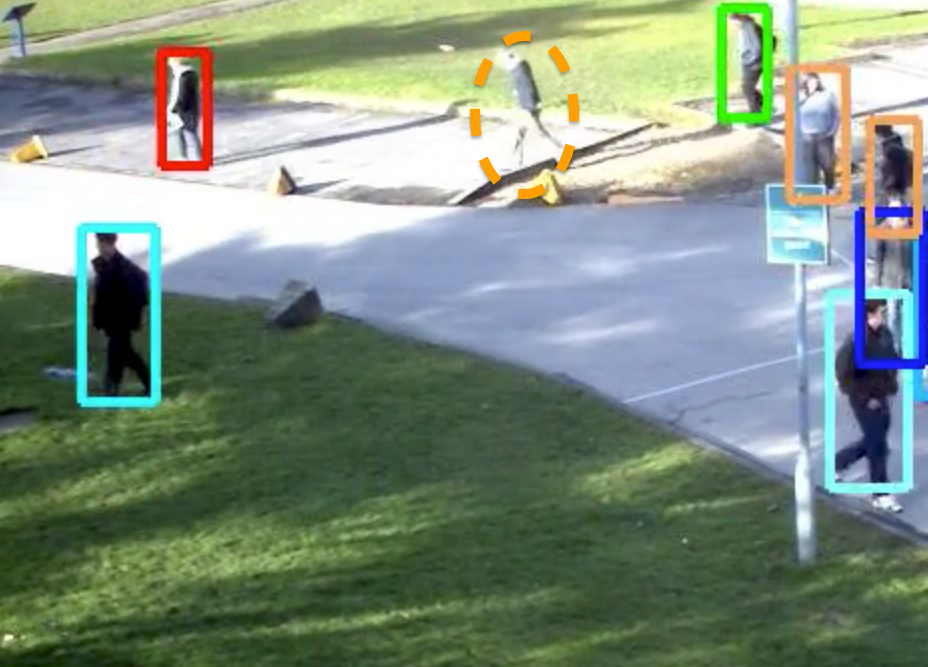}}
\vspace{-1.5mm}
\subfigure[Ours, frame 117]{\includegraphics[width = .3 \linewidth]{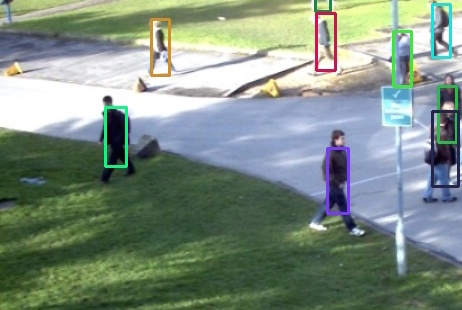}}
\subfigure[Ours, frame 125]{\includegraphics[width = .3 \linewidth]{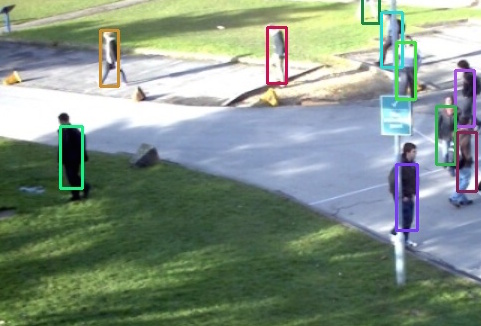}}
\subfigure[Ours, frame 128]{\includegraphics[width = .3 \linewidth]{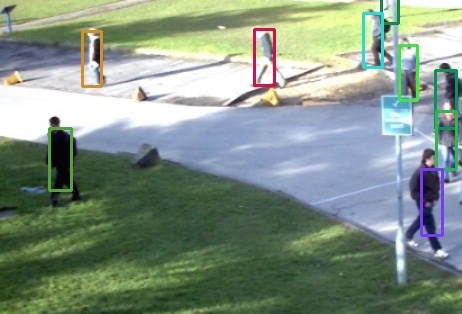}}
\end{center}
\caption{Qualitative evaluation: Results on 3dMOT 2015 PETS09-S2L2 sequence. Row 1-3 shows the tracking results of Milan et al. \cite{milan2016multi}, Milan et al. \cite{milan2017online} and Klinger et al. \cite{klinger2015probabilistic} respectively  whereas the fourth row presents the results of the proposed method. The dashed circles indicates the ID switches and missed pedestrians during the tracking}
\label{fig:qualitative_results}
\vspace{-3mm}
\end{figure}

A qualitative evaluation of the tracking results is shown in Fig. \ref{fig:qualitative_results}, where we compare the tracking outputs from the proposed method with state-of-the-art baselines. The ID switches and tracker misses are highlighted in dashed circles. It can be observed that Milan et al. \cite{milan2017online} performs poorly with frequent ID switches and missed tracks, as the tracker lacks neighbourhood information. The temporal modelling with probabilistic networks methods, Milan et al. \cite{milan2016multi} and Klinger et al. \cite{klinger2015probabilistic}, fail to anticipate the motion and generate erroneous tracking results.  In contrast, the proposed framework utilises the context and neighbourhood dynamics and generates accurate tracking results, even without using an appearance model.  

\begin{table}[]
\centering
\caption{Comparison of results from the proposed approach against state-of-the-art methods on ETHMS dataset. Arrows indicate favourable direction of each metric. Best values are printed in bold}
\label{tab:ETHMS}
\resizebox{.98\linewidth}{!}{
\begin{tabular}{|l|l|l|l|l|l|l|}
\hline
Tracker    & Recall $\uparrow $ & Precision $\uparrow $ & MT (\%) $\uparrow $ & ML (\%) $\downarrow$ & Frag  $\downarrow$& IDs $\downarrow$ \\ \hline
DP \cite{pirsiavash2011globally}        & 67.4   & \textbf{91.4}                           & 50.2 & 9.9 & 143  & 4   \\ \hline
PIRMPT  \cite{kuo2011does}   & 76.8   & 86.6                           & 58.4 & 8.0 & \textbf{23}   & 11  \\ \hline
Online CRF \cite{yang2012online} & 79.0   & 90.4                           & 68.0 & 7.2 & 19   & 11  \\ \hline
DCEM \cite{milan2016multi}      & 76.2   & 87.6                           & 58.3 & 7.1 & 78   & 43  \\ \hline \hline
\cellcolor[HTML]{C0C0C0}Proposed   & \cellcolor[HTML]{C0C0C0}\textbf{89.8}       & \cellcolor[HTML]{C0C0C0}91.0                               & \cellcolor[HTML]{C0C0C0}\textbf{73.8 }    &\cellcolor[HTML]{C0C0C0} \textbf{7.3}    & \cellcolor[HTML]{C0C0C0}25      &  \cellcolor[HTML]{C0C0C0}\textbf{3}   \\ \hline
\end{tabular}
}
\end{table}

We also evaluate the proposed tracking framework on the ETHMS dataset (see Tab. \ref{tab:ETHMS}) where a busy pedestrian street is filmed from a moving camera. We use pedestrian detections from the proposed detector and publicly available evaluation script. Methods like DP \cite{pirsiavash2011globally}, Online CRF \cite{yang2012online} and PIRMPT \cite{kuo2011does} are highly reliant on appearance based tracklet linking and occlusion avoidance. Yet our efficient data association scheme based on trajectory prediction outperforms these state-of-the-art methods with fewer ID switches and higher precision and recall. The DCEM \cite{milan2016multi} approach replaces appearance based data association with trajectory modelling, yet fails to generate accurate tracking results compared to the proposed method. 

\vspace{-2mm}
\section{Conclusion}
This paper has presented a new online deep learning framework for muli-person localisation and tracking. The proposed localisation framework builds upon generative adversarial networks and performs sequential modelling, allowing us to localise pedestrians in cluttered environments even in the presence of noise and other image artefacts. Our data association scheme utilises a trajectory modelling approach and anticipates human behavioural patterns under different contexts. It not only results in a light weight framework compared to other CNN based person re-identification architectures, but also introduces intelligence into the tracking framework via modelling different human behavioural patterns under different contexts such as group motion and random exploration.  Our evaluations on publicly available benchmarks have shown that the proposed method exhibits superior performance, especially among current state-of-the-art deep learning methods. The evaluations on both static and dynamic cameras ensures the applicability of the proposed method in variety of applications including autonomous driving, robotics, and egocentric vision. 

\vspace{-2mm}
\footnotesize{
\subsubsection*{Acknowledgement}
\vspace{-2mm}
This research was supported by the Australian Research Council's Linkage Project LP140100282 ``Improving Productivity and Efficiency of Australian Airports''. The authors also thank QUT High Performance Computing (HPC) for providing the computational resources for this research.
}

%-------------------------------------------------------------------------

{\small
\bibliographystyle{ieee}
\bibliography{egbib}
}

\end{document}